\documentclass[
]{ceurart}
\usepackage{listings}
\usepackage{bbm}
\usepackage{subcaption}
\usepackage{tikz}
\usepackage{todonotes}
\usepackage{hyperref}
\newcommand{\pprob}{\mathbb{P}}

\newcommand{\cvset}{\mathcal{C}}
\newcommand{\rawjobs}{\mathcal{J}}
\newcommand{\rawcv}{\mathcal{C}}
\begin{document}

\copyrightyear{2025}
\copyrightclause{Copyright for this paper by its authors. Use permitted under Creative Commons License Attribution 4.0 International (CC BY 4.0).}

\conference{AEQUITAS 2025: Workshop on Fairness and Bias in AI | co-located with ECAI 2025, Bologna, Italy}

\title{Causal Synthetic Data Generation in Recruitment}

\author[1]{Andrea Iommi}[%
orcid=0009-0007-8337-3695,
email=andrea.iommi@phd.unipi.it]
\cormark[1]

\author[1]{Antonio Mastropietro}[%
orcid=0000-0002-8823-0163,
email=antonio.matropietro@di.unipi.it]

\author[1,2]{Riccardo Guidotti}[%
orcid=0000-0002-2827-7613,
email=riccardo.guidotti@unipi.it]

\author[1,2]{Anna Monreale}[%
orcid=0000-0001-8541-0284,
email=anna.monreale@unipi.it]

\author[1,2]{Salvatore Ruggieri}[%
orcid=0000-0002-1917-6087,
email=salvatore.ruggieri@unipi.it]

\address[1]{University of Pisa, Italy}
\address[2]{ISTI-CNR Pisa, Italy}
\cortext[1]{Corresponding author.}

\begin{abstract}
The importance of Synthetic Data Generation (SDG) has increased significantly in domains where data quality is poor or access is limited due to privacy and regulatory constraints. 
One such domain is recruitment, where publicly available datasets are scarce due to the sensitive nature of information typically found in curricula vitae, such as gender, disability status, or age.
This lack of accessible, representative data presents a significant obstacle to the development of fair and transparent machine learning models, particularly ranking algorithms that require large volumes of data to effectively learn how to recommend candidates. 
In the absence of such data, these models are prone to poor generalisation and may fail to perform reliably in real-world scenarios.
Recent advances in Causal Generative Models (CGMs) offer a promising solution. 
CGMs enable the generation of synthetic datasets that preserve the underlying causal relationships within the data, providing greater control over fairness and interpretability in the data generation process.
In this study, we present a specialised SDG method involving two CGMs: one modelling job offers and the other modelling curricula. 
Each model is structured according to a causal graph informed by domain expertise. 
We use these models to generate synthetic datasets and evaluate the fairness of candidate rankings under controlled scenarios that introduce specific biases.
\end{abstract}


\maketitle

\section{Introduction}
Synthetic data generation is gaining importance, especially in contexts where data quality is low, privacy concerns are prominent, or regulatory constraints limit data availability. 
Poor-quality datasets, often containing missing or unrepresentative information, can significantly impair the performance of Machine Learning (ML) models~\cite{DBLP:Bauer_André}. 
In many domains, collecting real data is prohibitively expensive or logistically challenging, and ensuring coverage of all relevant scenarios is rarely straightforward~\cite{DBLP:AbufaddaM21}.
Moreover, in high-risk settings such as healthcare~\cite{DBLP:MurtazaAKMZB23}, business~\cite{DBLP:Assefa_Samuel}, or recruitment~\cite{DBLP:Beretta_Andrea}, extensive preprocessing and privacy-preserving measures often degrade data utility, further motivating the need for high-quality synthetic alternatives.
Synthetic Data Generators (SDGs) offer a promising solution to challenges related to data scarcity, privacy, and regulatory compliance~\cite{DBLP:Lu_Yingzhou}. 
In healthcare, for example, SDGs support disease modelling and drug discovery while preserving patient confidentiality. 
Models such as SynSys~\cite{DBLP:Dahmen_Cook} and CorGAN~\cite{DBLP:CorGAN} address data availability and privacy concerns in medical applications.
Similarly, in business domains, strict privacy regulations often hinder research and development. 
In~\cite{DBLP:Assefa_Samuel}, it is demonstrated how the SDGs can be utilised to simulate financial scenarios under specific constraints. 

As in other sensitive domains, the availability of accessible datasets in human recruitment is limited due to the private nature of attributes such as gender, disability, and age, which are pieces of information that candidates may be reluctant to disclose. 
Consequently, synthetic datasets play a crucial role in this field, enabling the training and evaluation of ranking models not only in terms of performance but also in terms of fairness within human recommendation systems. 
In addition to improving model effectiveness, synthetic data helps mitigate the risk of disclosing sensitive attributes, thereby addressing both ethical and legal concerns.
Unfortunately, generating synthetic datasets that accurately reflect real-world data is a non-trivial task. 
Indeed, to be effective, synthetic data must closely replicate the underlying statistical properties of the original data. 
Furthermore, in socially sensitive domains, the generation process must also ensure fairness and interpretability to prevent biased outcomes.

Causal Generative Models (CGMs) can address these needs by explicitly encoding causal relationships using Structural Causal Models (SCMs)~\cite{Peters_Jonas}.
Indeed, unlike deep learning approaches such as Generative Adversarial Networks (GANs)~\cite{DBLP:Goodfellow} or Variational Autoencoders (VAEs)~\cite{DBLP:autoencoder}, which excel at capturing complex non-linear patterns, CGMs provide transparent and interpretable mechanisms grounded in causality. 
While deep learning models can uncover correlations, they often fail to reveal the underlying causal structure and may introduce spurious associations due to opaque training processes~\cite{DBLP:Vallverdú_Jordi,DBLP:Komanduri_Aneesh}.
In fact, in high-risk domains, the importance of interpretability is underscored by the European Union’s AI Act, which mandates transparency and human oversight in AI systems~\cite{DBLP:Pavlidis}. 
This regulation highlights the need for transparent ML models and data generation processes that can be audited and understood by domain experts.

In this paper, we present a SDG system grounded on two CGMs, one for job offers and one for curricula, each structured according to causal graphs derived from interviews with domain experts. 
These graphs capture the decision-making processes underlying the creation of job offers and candidate profiles. We use the CGMs to generate synthetic datasets that simulate realistic recruitment scenarios. 

As a test-bed of our SDG, we explore fairness in ranking tasks.
We introduce a controlled bias by incorporating a parametric causal link between the \textit{gender} attribute and \textit{working hours}, simulating gender disparities as discussed in social sciences~\cite{Mazei}.
This setup enables us to assess how such bias, when propagated through the data, influences the fairness of rankings of job candidates produced by ML ranking models.
In summary, the contribution of this work is threefold: \emph{(i)} the formulation of causal graphs to model the HR domain for job offer and curriculum generation, \emph{(ii)} a new approach to a tabular synthetic data generation, causality-grounded and intrinsically interpretable and \emph{(iii)} a public and extendible GitHub Python repository\footnote{\url{https://github.com/jacons/CausalSDG}} in which are deployed Causal mechanisms that work with multiple data types.

The rest of this paper is organised as follows. 
After reviewing works on synthetic data generation in Section~\ref{sec:related}, we briefly review the key concepts behind our proposal in Section~\ref{sec:background}. 
Then, in Section~\ref{sec:methodology}, we describe our proposal. 
In Section~\ref{sec:experiments}, we present the experimental results. 
Finally, Section~\ref{sec:conclusion} summarises our contributions and outlines potential directions for future research.

\section{Related Work}
\label{sec:related}
We begin by reviewing the literature on synthetic data generation for tabular data.

Early statistical methods, such as SMOTE~\cite{DBLP:Chawla_Nitesh}, generate synthetic samples by fitting empirical distributions and interpolating between existing data points. While effective in leveraging marginal distributions, these techniques often fall short in capturing complex feature interactions, which can result in the generation of less realistic (low-fidelity) and less representative (biased) synthetic data.

Deep learning (DL)--based models, including Generative Adversarial Networks (GANs), Variational Autoencoders (VAEs), and Diffusion Models~\cite{DBLP:Lu_Yingzhou,DBLP:Bauer_André}, show impressive capabilities in learning complex data distributions. 
However, these models also present several limitations.
First, as model complexity increases, they require large volumes of training data. In data-scarce settings, these models often struggle to learn accurate distributions, resulting in poor generalisation.
Second, DL--based generative models are prone to \textit{mode collapse}. 
The latter is a phenomenon where the model generates samples from only a limited subset of the true distribution, namely, a subset with high probability mass. 
This issue, as discussed in~\cite{Kossale_Airaj}, undermines the diversity and representativeness of the synthetic data, making it difficult to ensure dataset quality.
Lastly, these models typically function as black boxes, offering limited transparency into the data generation process. 
This lack of interpretability poses challenges in high-stakes applications, where understanding the model’s behaviour is critical. 

In contrast to deep generative models, probabilistic graphical models, such as Bayesian Networks (BNs), are better suited under certain conditions, namely when \emph{(i)} datasets are limited in size, \emph{(ii)} consistency is critical, meaning generated samples must adhere to domain-specific constraints, and \emph{(iii)} strong correlations exist among features~\cite{DBLP:Schoen_Blanc,DBLP:Combrink_Herkulaas}.
BNs explicitly model the conditional dependencies between variables, allowing them to capture the joint distribution of the data more transparently. 
This structure enables BNs to replicate the statistical properties of an empirical dataset with greater interpretability compared to DL--based approaches. 
Structural Causal Models (SCMs) extend the capabilities of BNs by embedding causal semantics into the graphical structure. 
While BNs focus on statistical dependencies, SCMs incorporate structural equations that define how each variable is generated from its causes. 
This allows SCMs not only to model observational distributions but also to support interventional and counterfactual reasoning.
As a result, SCMs provide a more expressive framework for generating synthetic data that is both statistically consistent and causally grounded, making them particularly valuable in domains where fairness, transparency, and causal interpretability are essential.

A growing body of research has focused on developing methods for generating synthetic data that explicitly mitigate fairness concerns while preserving utility. 
Specifically, \cite{DBLP:Xu_Depeng,DBLP:Abroshan_Mahed,DBLP:DECAF} extend GAN--based methods, and \cite{DBLP:GenFair} adopts a genetic approach. 
Unlike these approaches, \emph{(i)} we focus on the specific domain of recruiting (curricula and job offer datasets), \emph{(ii)} we adopt a SCM approach with controllable bias parameters, and \emph{(iii)} the data generation process is fully interpretable.
The focus on a specific domain permits us to derive causal dependencies among features by eliciting them from expert knowledge, thus overcoming the difficulty of discovering causal dependencies from observational data. 
We instead use observational data to learn the structural equations given the known causal dependencies among features.
To the best of our knowledge, this is the first approach to using SCMs for SDG in the recruiting domain.

The closest works are \cite{DBLP:DECAF} and \cite{Barbierato_Enrico}.
\citet{DBLP:DECAF} is a generic approach, which assumes a given causal graph and learns the structural equations through conditional GANs.
The data generation process first intervenes on the derived SCM by eliminating dependencies that lead to unfairness in a downstream model (these dependencies may be specific to the fairness metric being considered).
Subsequently, it generates data by applying the (GAN--based) structural equations of the remaining dependencies. 
On the other hand, our method integrates fairness constraints directly into the causal mechanisms during the generative process, permitting us to produce fairness requirements without integrating post-hoc intervention.
\citet{Barbierato_Enrico} present a methodology for bias-controllable synthetic data generation using parametric causal mechanisms. 
Their experimental framework explores fairness metrics by systematically varying a bias parameter during the data generation process.
However, their approach is limited to continuous variables and lacks the ability to learn causal mechanisms directly from data, relying instead on predefined parametric forms. 
In contrast, our approach supports mixed data types and utilises a learned causal mechanism fitted from observational data.

\section{Background}
\label{sec:background}
To keep the paper self-contained, we provide a brief outline of the key concepts underlying our proposal.

\subsection{The ESCO Taxonomy and the EQF Classification System}
\label{sec:esco_eqf}
The \textit{European Skills, Competences, Qualifications, and Occupations} ~\cite{DBLP:Smedt_Johan} (ESCO, \href{https://esco.ec.europa.eu/}{https://esco.ec.europa.eu/}) is a multilingual classification system developed by the European Union (EU) to provide a standardised framework for describing skills and occupations. 
It aims to support better matching between individuals and job opportunities, as well as between education and labour market needs.

ESCO is organised into two main pillars: Occupations and Skills. 
The Occupations pillar provides a structured vocabulary for consistently describing occupations. 
This structure is based on hierarchical relationships between concepts, using the \textit{International Standard Classification of Occupations} (ISCO-08) as its foundational taxonomy. 
Each occupation entry includes several attributes, such as a Unique Resource Identifier (URI), a preferred term that represents the concept in a specific language, a set of non-preferred terms including synonyms, spelling variants, declensions, and abbreviations, and a textual description.
Similarly, the Skills pillar provides a taxonomy for describing competencies and knowledge. 
It mirrors the hierarchical organisation of the Occupations pillar and shares the same set of attributes. 
This consistency facilitates interoperability and integration across different systems and languages.
ESCO also defines associations between occupations and skills, categorising them as either essential or optional. 
These associations enhance the system’s ability to support detailed profiling and more accurate matching in both employment and educational contexts. 

The \textit{European Qualifications Framework} (EQF, \href{https://europass.europa.eu/en/europass-digital-tools/european-qualifications-framework}{https://europass.europa.eu/en/europass-digital-tools/european-qualifications-framework}) serves as a common European reference framework to facilitate the comparison of qualifications across different countries and education systems. 
Established by the EU, the EQF aims to promote transparency, mobility, and lifelong learning by aligning national qualification systems through a shared structure into eight levels: from Level 1, which corresponds to basic general knowledge and skills, to Level 8, which reflects the highest level of expertise, typically associated with doctoral-level qualifications. 
Each level is defined by a set of descriptors that express the expected learning outcomes in terms of knowledge, skills, and competence.

\subsection{Structural Causal Models}\label{sect:scm}

A Structural Causal Model (SCM)~\cite{DBLP:PearlCausality2009,Peters_Jonas,DBLP:journals/widm/NogueiraPRPG22} describes a data-generating process by relating random variables in cause-effect pairs. 
Let $\mathbf{X} = \{ X_1, \ldots, X_d\}$ be $d$ observable random variables, defined by a set $\mathbf{F}$ of structural equations: 
\begin{equation}
\label{eq:se}
X_i := f_i(\mathbf{PA}_i, U_i) \quad \quad \mbox{for}\ i = 1,\dots, d
\end{equation}
where $\mathbf{U} = \{ U_1, \ldots, U_d\}$ are $d$ independent exogenous (unobserved) random variables, and $\mathbf{PA}_i \subseteq \mathbf{X} \setminus \{X_i\}$ are the causal parents of $X_i$. The equations describe the causal mechanism by which an $X_i$ is generated from its causal parents and an exogenous variable $U_i$. 
Formally, a SCM $\mathcal{M}$ is a tuple $\mathcal{M} = \langle \mathbf{U}, P(\mathbf{U}), \mathbf{X}, \mathbf{F} \rangle$, where $P(\mathbf{U}) = \prod_i P(U_i)$ is the probability distribution of the exogenous variables. 
The parental relations in a SCM induce a \textit{causal graph} $\mathcal{G}$, in which the nodes represent random variables and a directed edge $X_j \to X_i$ denotes a causal relation between $X_j \in \mathbf{PA}_i$ and $X_i$.
We assume Directed Acyclic Graphs (DAGs), meaning there are no loops in $\mathcal{G}$, so the data generation process can proceed by following a topological order of the variables given the graph. 
Under the Markov property assumption, the induced probability on $\mathbf{X}$ can then be factorized as $P(\mathbf{X}) = \prod_i P(X_i|\mathbf{PA}_i, U_i)$.

Let us assume that we know the causal graph $\mathcal{G}$, and that we are given a dataset of i.i.d. observations. 
Parametric and non-parametric approaches can be used to infer the structural equations. 
In this work, we consider the following approach based on the type of $X_i$.

For a continuous variable $X_i$, we model the task as a regression problem of the dependent continuous variable $X_i$ given the independent variables $\mathbf{PA}_i$ and the exogenous variable $U_i$.
In particular, we assume additive noise, namely, the structural equation is of the form:
$X_i = f_i(\mathbf{PA}_i) + U_i$, where $f_i$ is a regressor trained from the dataset of observations, and the exogenous variable $U_i$ is assumed to be empirically distributed from the residual $X_i - f_i(\mathbf{PA}_i)$.

For $X_i$ discrete, we assume $X_i \sim Cat(\hat{f}_i(\mathbf{PA}_i))$, namely $X_i$ is categorically distributed with probabilities given by the predictions of a probabilistic ML classifier\footnote{We adopt a \textit{HistGradientBoostingClassifier} from the \textit{scikit-learn} Python library.} $\hat{f}_i$ trained from the dataset of observations. 
If a variable in $\mathbf{PA}_i$ is of set type (e.g., skills required in a job offer or those possessed by a job applicant), we first one-hot encode all possible values in the set.

For $X_i$ of set type with possible values in $\{v_1, \ldots, v_m\}$, then $f_i$ samples $n \sim U(n_{\mathit{min}}, n_{\mathit{max}})$ values without replacement, where each $v_j$ is sampled with probability equal to the empirical conditional distribution $P(v_j \in X_i | \mathbf{PA}_i)$ over the dataset of observations. Here, $n_{\mathit{min}}$ and $n_{\mathit{max}}$ are user parameters.

\section{Methodology}
\label{sec:methodology}
In this section, we outline: \emph{(i)} the formulation of causal graphs to model the HR domain for job offer and curriculum generation and \emph{(ii)} the design of a data preprocessing pipeline.

\subsection{Causal Graphs for SDG}
\label{sec:causal}
A central component of our approach is the causal graph representing the structure of job offers and curricula.
Rather than relying on automated causal discovery from observational data, we construct this graph through expert elicitation, based on qualitative insights gathered from HR professionals.
To this end, we conducted four semi-structured interviews with HR representatives from three Italian companies of varying sizes: one with over 50 employees, another with over $200$ employees, and a third with more than $5,000$ employees. 
These organisations operate in distinct sectors and organisational scales, providing a diverse perspective on hiring practices. 
The interviews were designed to uncover the real-world decision-making processes behind the creation of job offers.
Participants were asked to describe the key attributes considered when drafting job postings, as well as the relationships among these elements.
The insights obtained were instrumental in defining the structure and dependencies encoded in the causal graph, ensuring that it reflects domain-specific knowledge and practical constraints.

\begin{figure}
    \centering
    \includegraphics[width=0.60\linewidth]{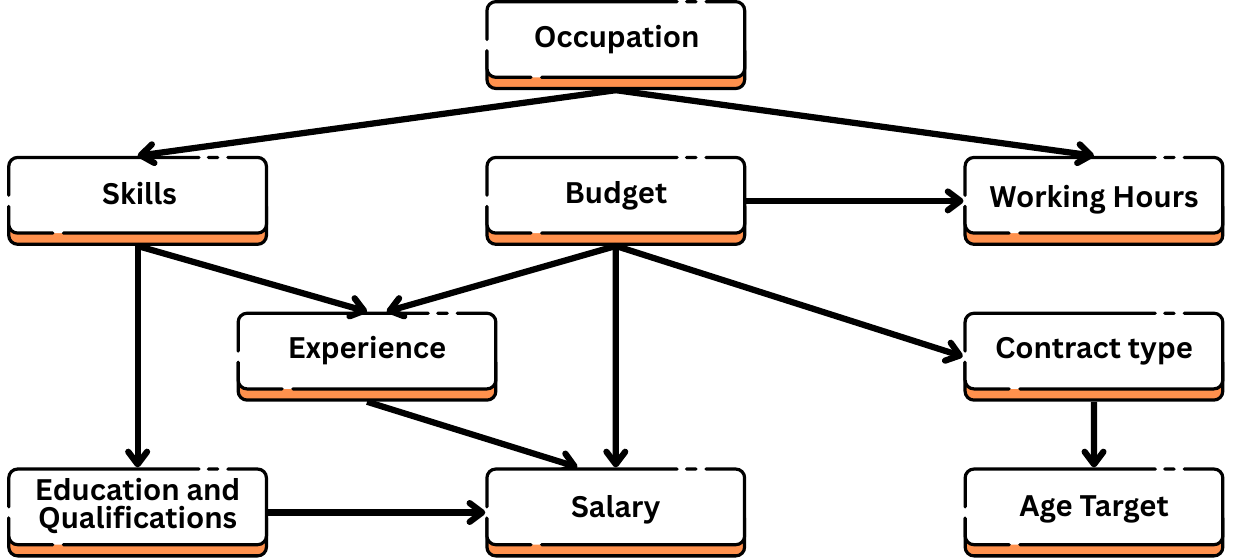}
    \caption{Job posting generation process as reconstructed from the interviews with HR professionals. The meaning of an arrow $A \rightarrow B$ is ``$A$ determines $B$''.}
    \label{fig:job_cf_interview}
\end{figure}

\paragraph{The Process Leading to Job Offers.}
\label{sec:aling_job_t}
In the companies interviewed, the recruitment process is closely integrated with annual budget planning.
Each production unit submits its needs for \textit{occupations}, which are then reviewed and approved through an internal administrative process. 
These needs may arise from various factors, including employee resignations, unplanned replacement costs, increased workloads, or skill shortages within teams.
A common sequence emerged across the interviews regarding how job postings are formulated. 
The process typically begins with the budget, which is defined during the annual planning phase. 
This \textit{budget} plays a pivotal role in shaping the job offer: it determines the experience of the sought candidate (e.g., junior or senior), the nature of the employment contract, and its duration. 
Specifically, higher budget availability, combined with organisational needs, affects the number of \textit{working hours} and the \textit{contract type} (intended as contract duration: permanent or fixed-term).
Under the guidance of the department head who initiated the staffing request, HR professionals identify the required hard skills.
Based on these, they determine the appropriate seniority level (e.g., years of \textit{experience}) and specify technical qualifications, such as degrees or certifications.
This structured approach reflects a rationale in which workforce needs are translated into a set of competencies that candidates must possess. 
These competencies are typically associated with specific educational backgrounds or levels of professional experience.
Moreover, defining education and experience through the lens of skills provides an initial validation mechanism: requiring a degree in a specific field, HR professionals implicitly assume that the candidate possesses the associated skills. Interestingly, some interviews revealed that the process can also operate in reverse—starting with an ideal candidate profile and subsequently adjusting it to fit within budget constraints.
Thus, the budget may either serve as the starting point for defining job requirements or act as a constraint to be considered after the ideal profile has been outlined.
Figure~\ref{fig:job_cf_interview} shows the job posting generation process as reconstructed from the interviews.

\begin{figure}
    \centering
    \includegraphics[width=0.8\linewidth]{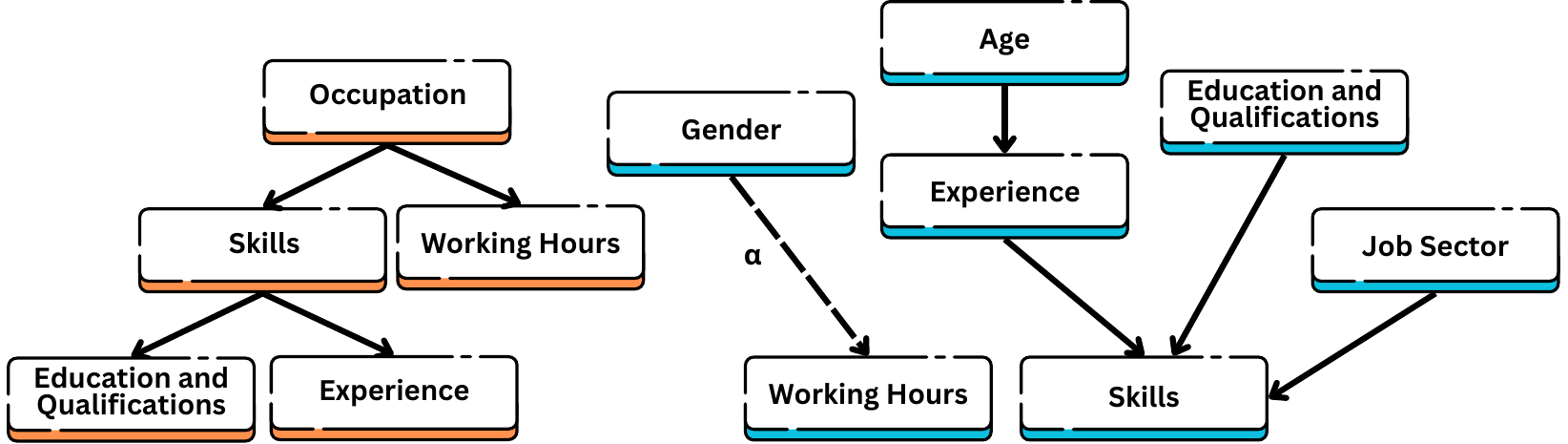}
    \caption{Causal graphs adopted for the experiments: left for job offers, right for curricula. The SDG developed is fully general and accepts DAG causal graphs as input. The dotted line in the curriculum's causal graph controls for bias through the parameter $\alpha$. In our experiment, all features are categorical/ordinal, except for ``skills'' which is a set of categorical values.
    }
    \label{fig:cf_x_experiment}
\end{figure} 

\paragraph{Causal Graphs for Job Offers and Curricula.}
\label{sec:cg}
Job offers typically do not include all of the variables of the graph from Figure~\ref{fig:job_cf_interview}. 
For instance, the budget of companies is not disclosed. 
Other variables may be missing for specific data collections. 
In our datasets (see next section), in particular, we lack the \textit{contract type}, the \textit{salary}, and the \textit{age target}.
To reflect these limitations, we present a simplified version of the causal graph used in our experiments, shown in Figure~\ref{fig:cf_x_experiment} (left). 
Despite this simplification, our SDG is designed to be fully parametric with respect to the input DAG--based causal graph. 
This design enables the system to be applied to richer datasets than the one used in our current experiments, granting broader applicability in future scenarios where more complete data is available.
The causal graph for job offers models the determination of required \textit{education} and \textit{experience} based on the \textit{skills} necessary for a given occupation. 
This reflects a form of backwards reasoning: starting from the desired skills, the HR professionals infer the \textit{qualifications} that provide a reasonable guarantee of possessing those skills. 
In other words, they answer the question: \textit{What educational background and years of experience are typically needed to ensure a candidate has the required competencies?}
Conversely, for job applicants, the causal direction is reversed.
A candidate’s \textit{skills} are shaped by their \textit{education} and years of \textit{experience}.
Additionally, the \textit{job sector} in which the candidate has specialised plays a significant role in skill development. 
It is important to note that \textit{occupation} is a more specific concept than \textit{job sector}. 
For example, while the \textit{job sector} might be \textit{Information and Communication Technology (ICT)}, the \textit{occupation} could be \textit{Python Developer}.
In the causal graph for curricula, \textit{age} is modelled as a determinant of the years of \textit{experience} a candidate has acquired. 
This reflects the natural progression of professional development over time. 
The full structure of this graph is shown in Figure~\ref{fig:cf_x_experiment} (right).

The graph also includes the variable \textit{working hours}, which represents whether a candidate is looking for a part-time or full-time contract. 
The study presented in~\cite{Mazei} offers valuable insights into the ways gender and societal norms influence working time patterns among men and women.
It highlights that women are significantly more likely than men to engage in part-time employment and are less inclined to work overtime.
Importantly, this trend is not merely a matter of personal preference but is deeply embedded in prevailing social constructs. 
The primary driver of this disparity lies in traditional gender roles, which continue to position women as the primary caregivers within the family. 
To model such a potential bias, we introduce a directed edge from \textit{gender} to \textit{working hours}, with the strength of this dependency modulated by a parameter $\alpha$. 
This parameter allows us to control the degree of gender--based bias in the data generation process and will be varied in the experimental analysis presented later in the paper. 
An example of a job offer and a corresponding curriculum, including the variables considered in their respective causal graphs, is provided in Table~\ref{tab:example_cv_job}.

\begin{table}[t]
   \caption{An example of a generated job offer (left) and a curriculum (right).}
    \centering
    \scriptsize
    \begin{tabular}[t]{>{\raggedright\arraybackslash}p{3cm}p{3.5cm}}
        \toprule
        Features & Values \\ \hline
        Occupation & \textit{ICT Professional} \\
        Working Hours & \textit{Full time contract} \\ 
        Education (EQF) & \textit{level 6} \\ 
        Experience & \textit{1 - 2 years} \\ 
        Skills & \textit{PHP, Java, French} \\
        \toprule
    \end{tabular}
    \quad
    \quad
    \scriptsize
    \begin{tabular}[t]{>{\raggedright\arraybackslash}p{3cm}p{3.5cm}}
        \toprule
        Features      & Values \\ \hline
        Job Sector    & \textit{ICT Professional}  \\
        Education (EQF)     & \textit{level 6} \\
        Gender        & \textit{Male} \\
        Working Hours & \textit{Full time contract} \\
        Age           & \textit{20} \\
        Experience    & \textit{2 years} \\
        Skills        & \textit{PHP, English, Groovy} \\
        \toprule
    \end{tabular}

    \label{tab:example_cv_job}
\end{table}

\subsection{Downstream Task: Ranking}
\label{sec:problem_setup}

The primary downstream application of the SDG framework is to evaluate, or potentially train, a ranking model that orders applicants for a given job position.
In this section, we present the end-to-end pipeline, from dataset generation to the evaluation of ranking performance and fairness metrics.
To enable the assessment of gender--based (un)fairness in ranking models, our SDG approach incorporates the parameter $\alpha$ within the causal graph generating curricula see Section~\ref{sec:cg}). 
This parameter allows for controlled interventions on gender-related attributes, facilitating rigorous fairness analysis in the ranking process.

\paragraph{Ranking Pipeline.}
Learning to Rank (LTR) methods originate from Information Retrieval~\cite{DBLP:Liu_Tie}. 
They are typically classified into three main categories w.r.t. a given user query: \textit{pointwise}, where individual documents are scored independently; \textit{pairwise}, where the model learns to predict the relative order between pairs of documents; and \textit{listwise}, where the entire list of documents is considered simultaneously to optimize a ranking metric.
In the context of recruitment, the job offer represents a user query, and the curricula identify the documents to be ranked.
Since candidates may choose which job offer to apply to, or at least to which job sector, the curricula to be ranked for a given job offer $j$ will be denoted by $\cvset_j$. 

Let us consider the case of pointwise ranking here.
A model is trained based on a dataset of observations consisting of pairs $(\mathbf{v}^{(j,c)}, s)$ where: $\mathbf{v}^{(j,c)}$ are fitness values of the curricula $c$ w.r.t. the job offer $j$, and $s$ is a relevance score assigned by an HR professional to the fitness values. 
The fitness values are numbers in the interval $[0, 1]$ that measure how much each requirement of a job offer is satisfied by a candidate, with $0$ indicating no match and $1$ indicating a full match. 
With reference to the graphs of Figure~\ref{fig:job_cf_interview}, a fitness value is computed for skills (resp., working hours/education and qualification/experience) of the candidate w.r.t. skills (resp., working hours/education and qualification/experience) of the job offer. 
For instance, the match for skills can be computed as the fraction of skills required by the job offer that the candidate possesses.
Regarding the relevance score $s$, when the data is collected from past candidate selections, it is provided by the HR professional assessing the candidates. 
In the case of synthetically generated data, such information is lacking. 
To experiment with our SDG, we assume a linear model:
\begin{equation}
    \label{formula:ranking_model}
    s{(j,c)} = \boldsymbol{w}^T\boldsymbol{v}^{(j,c)} + \beta
\end{equation} 
where $\boldsymbol{w} \in \mathbb{R}^r$ is a weight vector that encodes the importance of each fitness value and $\beta \sim N(0, \sigma^2)$ is a small noise value modelling uncertainty. 
Due to their interpretability, linear models are often adopted in real ranking systems.
Eq.~(\ref{formula:ranking_model}) provides the ``ground-truth'' for evaluating a learned ranking model.
We emphasise that, while an actual recruitment system would utilise a ranker such as LambdaMART or ListNet with scores assigned by HR, training and evaluating such models is beyond the scope of this work.
Indeed, in our experiment, we are interested in how fairness measurement changes at varying values of $\alpha$'s parameters. 

For a fixed job offer $j$, the ground-truth ranking of curricula in $\cvset_j$ is the one obtained by descending value of $s{(j,c)}$. 
Candidates with the same score will have the same ranking position. 
Let us denote by $\pi^j_c$ the rank position of $c$. 
We further denote by $\hat{\pi}^j_c$ the rank position of $c$ as determined by another (e.g., a learned) ranking model.

\paragraph{Performance and Fairness Metrics.}
Several evaluation metrics have been considered to quantify the performance of ranking models. 
They measure how close $\hat{\pi}^j_c$  and $\pi^j_c$ are over all ranked candidates (Kendall's tau, Spearman's rho) or over the top $k$ ranked candidates (precision, NDCG), for a fixed job offer $j$, or for a set of job offers (average precision). 
In this paper, however, we are not evaluating ranking models, but rather the properties of the datasets generated by our SDG.
The performance metric of a SGD is the fidelity of the generated data to the distribution of real data.
However, due to the issues mentioned regarding the collection of representative real data, we opt for a ``by-design'' approach, assuming that the causal graphs depicted in Figure~\ref{fig:cf_x_experiment} are faithful to the true distribution of job offers and curricula. 
Conditional distribution of a variable given the parent nodes is enforced by fitting the structural equations as described in Section~\ref{sect:scm}. 
Whether or not the causal graphs from Figure~\ref{fig:cf_x_experiment} are valid in a specific context, such as countries or industry sectors, remains to be determined context by context, possibly adapting the causal graphs in case of different/additional variables that are observable.

The other property that our SDG can model is the bias of the generated data.
We link such a notion to the (un)fairness of the downstream task.
Several fairness metrics have been considered in the literature~\cite{DBLP:Pitoura,DBLP:journals/csur/ZehlikeYS23a}.
Since we do not consider a ranking model learned from data, we have to restrict ourselves only to metrics that regard the ground-truth ranking model from Eq.~\ref{formula:ranking_model}, possibly at the variation of the weights $\boldsymbol{w}$. We will consider the group fairness metrics~\cite{DBLP:Pitoura} of demographic parity (DP) and normalised discounted difference (rND):
\begin{equation}
   DP(j) = 1 - (\pprob(\pi^j_c \leq k |\  c\ \textit{protected}, c \in \cvset_j) - \pprob(\pi^j_c \leq k |\ c\ \textit{unprotected}, c \in \cvset_j ))    
\end{equation}

$DP(j)$ ranges in $[0,2]$, with $2$ denoting full unfairness against the protected group (only candidates from the protected group are selected in the top $k$ positions), $1$ denoting fairness (equal probability of being chosen among protected and unprotected), and $0$ denotes reverse unfairness against the unprotected group. 
The rND metric improves over DP by considering: (1) the deviation of the fraction of protected candidates in top-$i$ position against the proportion of protected candidates applying for the position; (2) at multiple thresholds $i$'s.
%
\begin{equation}\label{eq:rnd}
    rND(j) = \frac{1}{Z} \sum_{i=5, 10,15,\dots}^{|\cvset_j|}\frac{1}{log_2(i)}  | \frac{|\{\pi^j_c \leq k, c\ \textit{protected}, c \in \cvset_j\}|}{|\{\pi^j_c \leq k\}|} - \frac{|\{c\ \textit{protected}, c \in \cvset_j\}|}{|\cvset_j|}|
\end{equation}
where $Z$ is a normalising factor (making it comparable across different $j$'s) so that $rND(j) \in [0, 1]$, where $0$ means fairness (proportional representation of protected candidates in top positions).

\paragraph{Bias Parametrization.}
\label{sec:parametrized_alpha}
We describe here how the parameter $\alpha$ of the causal graph of the curricula (Figure~\ref{fig:cf_x_experiment} right) can be used to control the degree of bias induced in the generated datasets. 
Let us denote the variables \textit{gender} and \textit{working hours} as $G$ and $W$. 
$G$ is the sensitive variable, assuming the value $0$ as \textit{male} (unprotected group) or $1$ for \textit{not male} (protected group), while $W$ assume values $f$ for \textit{full-time} or $p$ for \textit{part-time}.
When there is no edge between $G$ and $W$, the generation of working hours boils down to the empirical distribution $P(W \mid G)$ over the dataset of observations. 
The parameter $\boldsymbol{\alpha}$, which we assume to be a pair $\boldsymbol{\alpha} = (\alpha_0, \alpha_1)$, can then be used to shift such a distribution either for the unprotected group, through $\alpha_0$, or for the protected group, through $\alpha_1$. 
Exponential tilting, power transformation, or other probability-shifting/skewing methods can be used. 
We adopt here exponential tilting, for which the shifted conditional probabilities are: $P'(W=p \mid G=0) = \, e^{-2 \alpha_1}/C_1 \cdot P(W = p \mid G=0)$, and $P'(W=f \mid G=0) = \, e^{-2 \alpha_1}/C_1 \cdot P(W=f \mid G=0)$, where $C_1$ is a normalizing constant. 
Similarly, we shift the conditional probability of the protected group: 
$P'(W=p \mid G=1)=\, e^{- 2\alpha_0}/C_0 \cdot P(W=p\mid G=1)$, and 
$P'(W=f\mid G=1)= \,e^{-2 \alpha_0}/C_0 \cdot P(W=f \mid G=1)$,
where $C_0$ is a normalising constant.

\section{Experiments}
\label{sec:experiments}

\begin{table}[t]
\caption{The distribution of \textit{working hours} conditional to \textit{gender} in the synthetic curricula datasets at the variation of $\alpha = (\alpha_0, \alpha_1)$. ``Part'' stands for  part-time contract, and ``Full'' for full-time contract.}
\centering
\scriptsize
\begin{tabular}{lrrrrrrrrrr}
    \toprule
     & \multicolumn{2}{c}{$\alpha_0 = -4$} & \multicolumn{2}{c}{$\alpha_0 = -1.5$} & \multicolumn{2}{c}{$\alpha_0 = 0$} & \multicolumn{2}{c}{$\alpha_0 = 1.5$} & \multicolumn{2}{c}{$\alpha_0 = 4$} \\ 
    & \multicolumn{1}{c}{Part} & \multicolumn{1}{c}{Full} & \multicolumn{1}{c}{Part} & \multicolumn{1}{c}{Full} & \multicolumn{1}{c}{Part} & \multicolumn{1}{c}{Full} & \multicolumn{1}{c}{Part} & \multicolumn{1}{c}{Full} & \multicolumn{1}{c}{Part} & \multicolumn{1}{c}{Full} \\ \hline
    
    male & 0.24 & 0.76 & 0.24 & 0.76 & 0.24 & 0.76 & 0.24 & 0.76 & 0.24 & 0.76 \\
    not-male & 0.02 & 0.98 & 0.25 & 0.74 & 0.59 & 0.41 & 0.87 & 0.13 & 0.99 & 0.01 \\ \hline\hline

    & \multicolumn{2}{c}{$\alpha_1 = -4$} & \multicolumn{2}{c}{$\alpha_1 = -1.5$} & \multicolumn{2}{c}{$\alpha_1 = 0$} & \multicolumn{2}{c}{$\alpha_1 = 1.5$} & \multicolumn{2}{c}{$\alpha_1 = 4$} \\ 
    & \multicolumn{1}{c}{Part} & \multicolumn{1}{c}{Full} & \multicolumn{1}{c}{Part} & \multicolumn{1}{c}{Full} & \multicolumn{1}{c}{Part} & \multicolumn{1}{c}{Full} & \multicolumn{1}{c}{Part} & \multicolumn{1}{c}{Full} & \multicolumn{1}{c}{Part} & \multicolumn{1}{c}{Full} \\ \hline
    male & 0.01 & 0.99 & 0.06 & 0.94 & 0.24 & 0.76 & 0.58 & 0.42 & 0.95 & 0.05 \\ 
    not-male & 0.60 & 0.40 & 0.60 & 0.40 & 0.59 & 0.41 & 0.60 & 0.40 & 0.61 & 0.39 \\      
    \toprule
\end{tabular}

\label{tab:distr_work_gender}
\end{table}
     
\paragraph{Experimental Settings.}
In this section, we illustrate some experiments on the functionalities of our SDG. 
For each experimental parameter ($\alpha_0$ and $\alpha_1$, which will be discussed later), we conducted $10$ runs, where in each one we generated $300$ job offers and $1000$ curricula.
For each job offer, all curricula are considered as candidates.
The motivation behind the generated data size is that, in our ranking pipeline, there is no effective model training process, except for the generator. 
In particular, the relevance score for each candidate is calculated using Eq.\ref{formula:ranking_model}, and the ranking is produced accordingly. 
Hence, the number of job offers and curricula generated for each run represents a good compromise between observing the changes in metrics and the execution time.

Data is generated according to the causal graphs shown in Figure~\ref{fig:cf_x_experiment}, with the structural equation fitted from the preprocessed datasets described in Appendix~\ref{sec:dataprep}.
The share of male and not-male (women and non-binary) applicants is $50\%$-$50\%$.
Regarding the ranking model pipeline, the fitness values between curriculum and job offer characteristics are defined through matching functions (MFs).
In particular, the MF for \textit{education} returns $1$ if the candidate’s education level is equal to or greater than the level required in the job offer; otherwise, it is $0$.
The MF for \textit{experience} checks whether the candidate’s experience falls within the interval required by the job offer. 
The MF for \textit{skills} calculates the fraction of skills required by the job offer that the candidate possesses.
Finally, the MF for \textit{working hours} tests for equality of the form required in the job offer and the form desired by the candidate. For example, for the data from Table~\ref{tab:example_cv_job}, the fitness value vector is $\boldsymbol{v}^{(j,c)} = [1.0,1.0,1/3,1.0]$ respectively for the MF of the features \textit{education, experience, skills} and \textit{working hours}.

In the following experiments, we investigate how changing the \textit{working hours} preferences conditional on \textit{gender} affects the fairness metrics DP and rND.
These kinds of experiments aim to simulate social norms that often say that women should prioritise family responsibilities over their careers, leading them to opt for part-time contracts that offer more free time but may limit career advancement opportunities.
The generated job offers exhibit a skewed distribution of working hours: $86.6\%$ of the positions are full-time contracts, while only $13.3\%$ are part-time contracts.
Also, the generated curricula have skewed distributions. Male applicants prefer full-time contracts ($76\%$), while not-male applicants prefer part-time contracts ($59\%$).
We will be varying $\alpha = (\alpha_0, \alpha_1)$ to investigate shifted distributions.
Table~\ref{tab:distr_work_gender} presents the average shifted distributions over the 10 experimental runs for a few $\alpha_0$ and $\alpha_1$ (see Section \ref{sec:parametrized_alpha}). 
In the experiments, we will be varying $\alpha_0$ and $\alpha_1$ from $-4.0$ to $4.0$.

In particular, we denote by $\alpha_0$  the parameter for the ``No-man'' distribution and by $\alpha_1$ the one for ``Man''.
We recall that larger $\alpha$'s result in a redistribution of probability mass towards part-time.
Notice that for $\alpha_0 = -1.5$, the distributions of male and not-male are almost identical. The same occurs for $\alpha_1 = 1.5$.

We explore four weighting vectors for the ranking model of Eq. (\ref{formula:ranking_model}).
All of them assign the same weights to the fitness values of \textit{education} (0.8), \textit{experience} (0.5), and \textit{skills} (1.0), and set $\beta \sim N(0, 0.01^2)$, while for \textit{working hours} we consider four cases: $0$, $0.5$, $0.8$ and $1.0$. 
Since \textit{gender} can only affect the score through \textit{working hours} (cfr. Figure~\ref{fig:cf_x_experiment} right), when the weight is $0$, then the scores of the ranking model are independent of the \textit{gender}, hence the ranking model is fair.

We emphasise that the weights considered for the experiment correspond to a ``rough approximation'' of the general HR decision-making.
Whether or not they constitute the effective importance value in a candidate evaluation is beyond the scope of this work. 
What matters is evaluating the fairness metrics at varying \textit{working hours} importances and the $\alpha$'s values.

\begin{figure}
    \centering
    
    \begin{subfigure}{0.46\textwidth}
        \includegraphics[width=\linewidth]{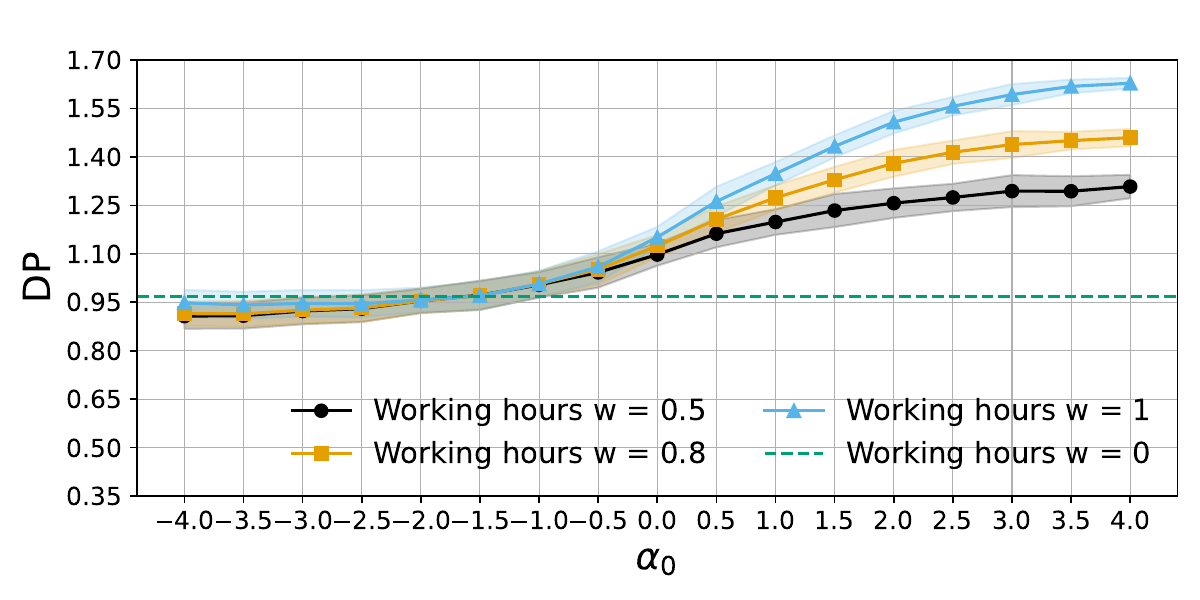}
        \caption{}
        \label{fig:dp_move_noman}
    \end{subfigure}\hfill
    \begin{subfigure}{0.46\textwidth}
        \includegraphics[width=\linewidth]{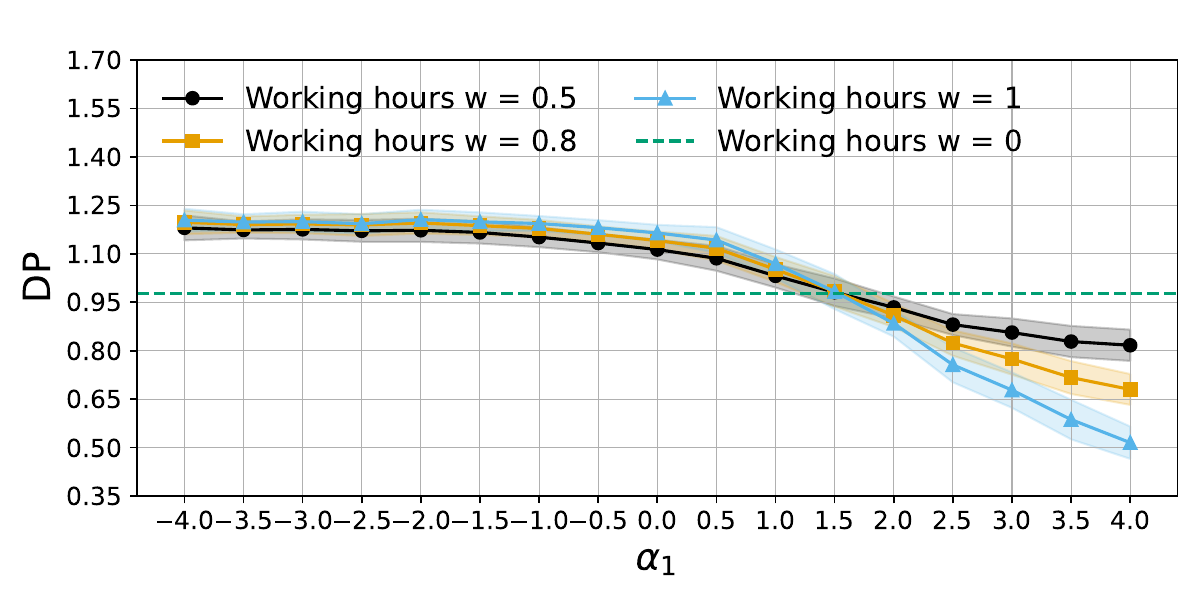}
        \caption{}
        \label{fig:dp_move_man}
    \end{subfigure}
    \begin{subfigure}{0.46\textwidth}
        \includegraphics[width=\linewidth]{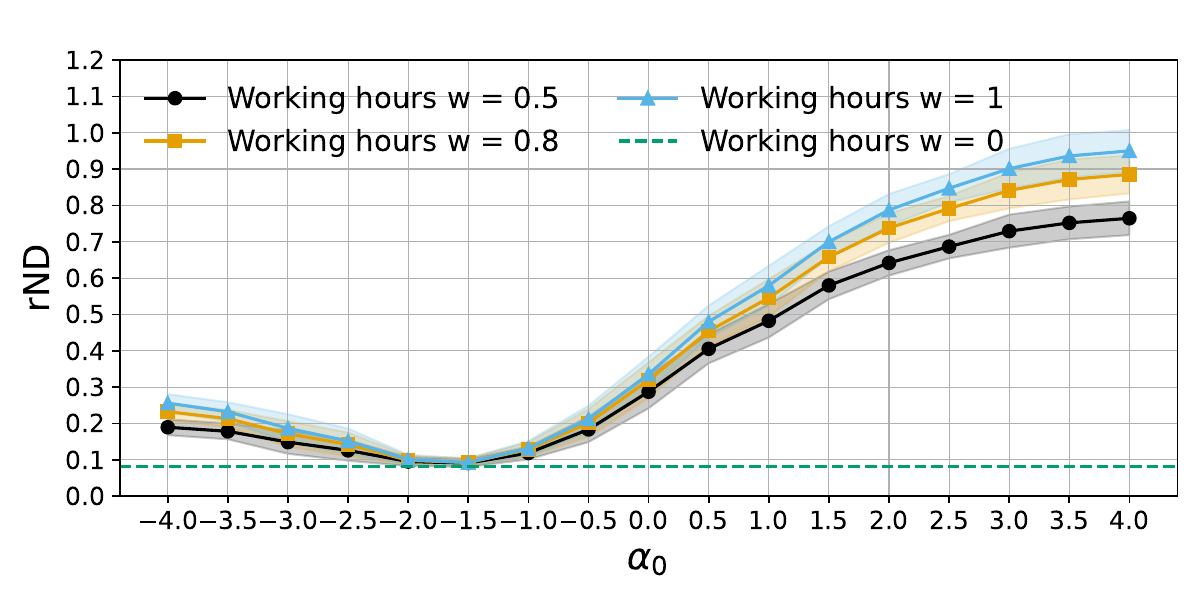}
        \caption{}
        \label{fig:rnd_move_noman}
    \end{subfigure}\hfill
    \begin{subfigure}{0.46\textwidth}
        \includegraphics[width=\linewidth]{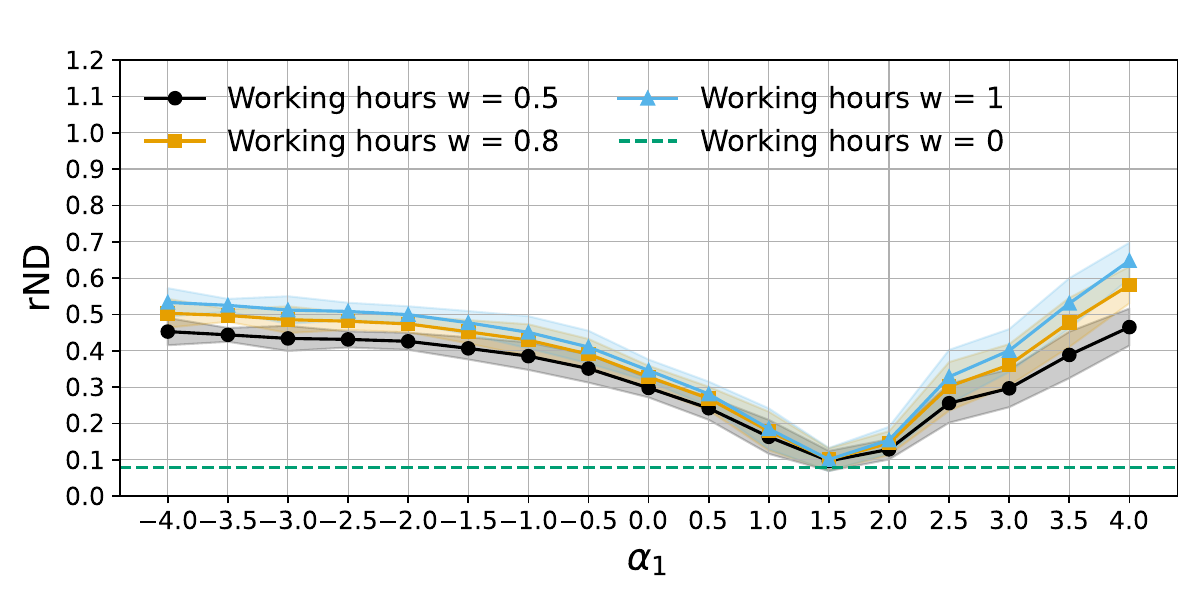}
        \caption{}
        \label{fig:rnd_move_man}
    \end{subfigure}

    \caption{Demographic Parity difference (DP) and normalised Discounted Difference (rND) at the variation of the bias-controlling parameters $\alpha_0$ (for not-male) and $\alpha_1$ (for male). DP and rND are averaged over 10 runs, each one with 300 job offers. The series represents the ranking model of Eq. (\ref{formula:ranking_model}) for different weights to the fitness value of \textit{working hours}. Shadows represent $\pm 1$ standard deviation over the $10$ runs. }
    \label{fig:dp_rnd}
\end{figure}
 
\paragraph{Results.} 
Figure~\ref{fig:dp_move_noman} illustrates the behaviour of Demographic Parity (DP) under the not-male conditional distribution shift. 
When the parameter $\alpha_0$ is negative, the distribution of preferences for not-male becomes skewed towards full-time employment. 
This shift results in a more balanced alignment between job requirements and candidate preferences. 
Hence, DP is close to the fair value of $1$ for all weights of \textit{working hours} in the ranking model.
However, as $\alpha_0$ increases, the distribution shifts toward part-time contracts.
Given the dominance of full-time job offers, this misalignment causes the ranking system to increasingly favour candidates from the non-protected group, i.e., males. 
As a result, DP values decline significantly, highlighting a growing disparity and reduced fairness for the protected group, except for the ranking model, where the mediating variable \textit{working hours} has zero weight.

A reversed trend is observed in Figure~\ref{fig:dp_move_man}, which illustrates the variation of DP with respect to changes in the parameter $\alpha_1$. 
For negative values of $\alpha_1$, male candidates are predominantly associated with full-time contracts. 
As a result, they are favoured by the ranking model (except in the case where the model assigns zero weight to the \textit{working hours} feature), leading to high DP values that indicate potential unfairness against non-male candidates.
As $\alpha_1$ increases, the distribution of male candidates shifts toward part-time contract preferences.
This shift reduces their alignment with the majority of job offers, which are primarily full-time, and causes the model to assign higher scores to non-male candidates. 
Hence, DP values fall below 1, highlighting unfairness against male candidates.
Notice that fairness (DP $\approx 1$) is achieved for all ranking models when the conditional distributions of \textit{working hours} are the same for male and not-male, namely for $\alpha_1=1.5$ (cfr. also Table \ref{tab:distr_work_gender}).

Similar patterns can be observed for the rND metric in Figures~\ref{fig:rnd_move_noman} and~\ref{fig:rnd_move_man}. 
The metric is computed by considering $i=5, 10, 15, 20$ in Eq. (\ref{eq:rnd}). Unlike Demographic Parity (DP), which measures average differences in exposure between groups, rND captures absolute deviations. 
As a result, trends where DP falls below 1 (indicating underexposure of the protected group) correspond to elevated rND values. 
This reflects a greater degree of ranking disparity, as rND penalises any deviation from perfect parity, regardless of direction.
As with DP, the most favourable rND values are observed at $\alpha_0=-1.5$ and $\alpha_1=1.5$, corresponding to the settings where the conditional distributions of working hours are equivalent for male and non-male candidates.
These configurations yield the most balanced exposure across groups, minimising both average and absolute disparities in ranking outcomes.

\section{Conclusions}
\label{sec:conclusion}
We introduced a synthetic data generation system for recruitment, based on expert-informed causal graphs that enable explicit control over bias and interpretability.
Through causal modelling, we assessed the fairness impact of biased data on a linear point-wise ranking model using DP and rND metrics.
Our experiments demonstrate that controlled distributional shifts in the generative process can significantly influence ranking fairness—positively or negatively—especially as the weight of bias-related features increases.
To foster further research in high-risk human recommendation scenarios, we will release the system as open-source software\footnote{\url{https://github.com/jacons/CausalSGD}}.

Despite promising results, two main limitations remain: the current approach supports only tabular data and depends on the availability of both high-quality training data and well-structured causal graphs; additionally, our experiments suffered from heterogeneous and non-standardised training data, which required approximations during preprocessing and may have reduced representativeness.
Future work will focus on \emph{(i)} make comparisons with closest related works pointing out advantages and limitations between the generative methods, \emph{(ii)} enhancing feature engineering to expand causal graphs with additional attributes, \emph{(iii)} analysing more complex discrimination scenarios like the intersectional bias, and \emph{(iv)} integrating into the SDG external knowledge, such as ESCO ontology, to improve the data diversity in the generated data.

\paragraph{Use of generative tools}

During the preparation of this work, I used Grammarly in order to improve my grammar and
spelling checking. 
After using these tools, I reviewed and edited the content as needed.

\bibliography{bibl}

\appendix

\section{Data Preprocessing for SDG}
\label{sec:dataprep}
The other key input of SDG consists of two datasets of i.i.d. observations of job offers and curricula.
From such a dataset, we can derive the structural equations of the SCM as described in Section~\ref{sect:scm}.
However, several preprocessing steps are necessary to ensure the datasets are of high quality and suitable for modelling. 
To facilitate this, our system provides a suite of APIs designed to perform various data transformations aimed at standardising and enriching the raw data. 
We take advantage of a Large Language Model (LLM)\footnote{Specifically, we use \textit{gemma2-9B}, available at \href{https://huggingface.co/google/gemma-2-9b}{https://huggingface.co/google/gemma-2-9b}.} to align the raw data with the ESCO taxonomy (see Section~\ref{sec:esco_eqf}). 
The task of extracting features from curricula and job offers through LLMs is receiving increasing attention. Recent studies have addressed similar challenges, as illustrated in works such as \cite{DBLP:journals/dke/PalshikarPBSRPTBJHCMC23,DBLP:conf/nldb/GanM23}.
It is important to emphasise that while the ESCO and EQF are EU-centric taxonomies, they serve as tools to ensure comparability between job offers and curricula. 
Our proposed methodology is modular, incorporating external knowledge, such as the O*NET ontology adopted in the US.

In this section, we describe the preprocessing procedures applied to two raw data sources: a real-world dataset of 10,000 job offers ($\rawjobs$) and a semi-synthetic dataset of 1,020 curricula ($\rawcv$). 
The job offers dataset was provided by a major recruiting company based in Spain.
The curricula dataset was obtained by the FINDHR project (\href{https://github.com/findhr}{https://github.com/findhr}), which combines features extracted from real-world curricula voluntarily donated for research purposes. 
Several challenges arise when working with these two data sources. 
First, the job offers are written in Spanish, whereas the curricula are in English. 
Second, the underlying taxonomies for education, qualifications, and related attributes differ significantly between the two datasets.
The following subsections detail the preprocessing transformations applied to harmonise these datasets, organised by the features represented in the two DAGs shown in Figure~\ref{fig:cf_x_experiment}.
\paragraph{Preprocessing Education and Qualifications.}
Education and qualification naming conventions vary significantly across countries and may also differ between the two datasets, $\rawjobs$ and $\rawcv$. 
For instance, in Spain, educational credentials are typically expressed using the national system, including degrees such as ``\textit{Bachillerato}'', ``\textit{Ingeniería Técnica}'', and ``\textit{Ciclo Formativo de Grado Superior}''. 
In contrast, English-language curricula often refer to broader categories such as ``\textit{Bachelor’s Degree}'' or ``\textit{Master’s Degree}'' to denote tertiary education levels.
To enable meaningful cross-country comparisons, we adopted a mapping developed by domain experts to translate raw educational and qualification data into the European Qualifications Framework (EQF) levels (see Section~\ref{sec:esco_eqf}). 
For example, ``\textit{Bachillerato}'' corresponds to EQF Level 4, while ``\textit{Ingeniería Técnica}'' maps to EQF Level 6, which is generally equivalent to a ``\textit{Bachelor's degree}'' in other national systems.

Job offer descriptions typically specify a single education level, usually indicating the minimum required qualification for the position.
In contrast, curricula often list multiple educational achievements. 
Thus, determining a candidate’s EQF level in the $\rawcv$ dataset involves several steps.
We begin by applying a keyword--based approach to extract and classify educational entries. 
For example, a phrase such as ``BSc in Computer Science'' is interpreted as a Bachelor's degree.
These inferred degree types are then mapped to their corresponding EQF levels. 
When a candidate's CV includes multiple qualifications, we select the highest EQF level as a representative indicator of their overall educational attainment.

\paragraph{Preprocessing Occupation and Job Sector.}
In the $\rawjobs$ dataset, there is a ``job title'' feature that captures the title associated with each job offer.
However, these titles exhibit considerable variability in both structure and specificity.
For example, some entries are well-defined, such as ``Data Scientist'' or ``ICT System Architect'', while others are either overly generic (e.g., ``Developer'') or excessively detailed (e.g., ``Web Developer (PHP, JS proficiency) – full remote contract''). 
This inconsistency often stems from differing practices among HR departments. 
Some prefer concise titles with detailed descriptions, while others embed extensive information directly into the job title field.
Such heterogeneity complicates the task of comparing or categorising job offers in a standardised manner.
To address this, we implemented a multi-phase alignment process that leverages LLMs and the ESCO taxonomy to normalise job titles across the dataset.
Such an alignment process consists of three steps.
\textit{Step 1: Title Refinement via LLM}. We use an LLM to generate a cleaner, more representative job title based on the original title and the accompanying job description. This step reduces noise and ambiguity, making titles that better reflect the underlying occupation.
\textit{Step 2: ESCO Occupation Retrieval.} We query the ESCO API (\href{https://esco.ec.europa.eu/en/use-esco/use-esco-services-api}{https://esco.ec.europa.eu/en/use-esco/use-esco-services-api}) using the LLM-generated job title. The API returns a list of relevant ESCO occupations, which serve as candidate labels for standardisation.
This step is crucial, as the ESCO search engine performs more effectively when provided with well-structured input.
\textit{Step 3: Final Classification via LLM}. We use the LLM again to select the most appropriate ESCO occupation label from the list retrieved in the previous step. This ensures that each job title (i.e., \textit{occupation} in the terminology of Figure~\ref{fig:cf_x_experiment}) is mapped to a standardised occupational category.

For the $\rawcv$ dataset, job sectors were already relatively normalised. Therefore, we applied only the second and third steps of the alignment process: querying the ESCO API and classifying the result using the LLM.
Through the above pipeline, we achieved a consistent ESCO--based standardisation of \textit{occupation} and \textit{job sector} across both datasets, enabling reliable comparisons and downstream modelling.

\begin{figure}
    \centering
    \includegraphics[width=0.8\linewidth]{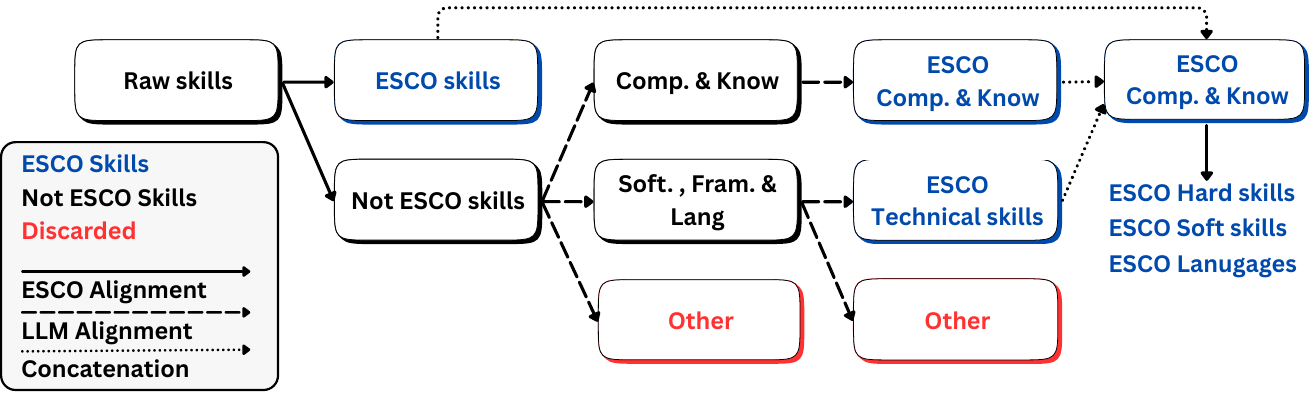}
    \caption{Preprocessing pipeline for skills.}
    \label{fig:skill_preprocessing}
\end{figure}

\paragraph{Preprocessing Skills.}
The skill domains in the two datasets differ significantly, making direct comparison non-trivial. 
The objective of the skills alignment process is to produce an ESCO-compliant list of skills for both $\rawjobs$ and $\rawcv$.
To achieve this, we design a multi-step procedure that leverages contextual information, specifically, the ESCO \textit{occupation} in $\rawjobs$ or the ESCO \textit{job sector} in $\rawcv$, previously aligned (see previous Section \ref{sec:aling_job_t}).

The \textit{first step} involves separating skills that are already ESCO-compliant from those that are not. 
This is accomplished by matching skill terms against the ESCO taxonomy depending on the language of the dataset. 
Then, we search the terms among the principal concepts, known as preferred labels, and the associated synonyms provided by the ESCO APIs.

In the \textit{second step}, we focus on the non-ESCO skills. These are further classified into three categories using the LLM: ``\textit{Competence and Knowledge}'', ``\textit{Software, Frameworks, Tools and Similar}'', and ``\textit{Other}''. 
The classification is performed by using an in-context learning approach, where the LLM receives not only the skill to be classified but also contextual information such as the job sector and the surrounding skill set.
This context significantly improves classification accuracy by allowing the model to consider semantic relationships beyond isolated terms.
For skills categorised as ``\textit{Competence and Knowledge}'', we query the ESCO API to retrieve a list of relevant ESCO skill terms.
The LLM is then used to select the most semantically appropriate term based on the provided context. For ``\textit{Software, Frameworks, Tools and Similar}'', we first manually selected a set of ten representative ESCO concepts. 
The LLM then associates each technical skill with the most relevant concept from this list. 
This step was necessary because the ESCO API often fails to return meaningful results for highly specialised technical terms not covered by the taxonomy.
Skills classified as ``\textit{Other}'' were deemed too noisy or incomplete by the LLM and were excluded from further processing.

In the \textit{final step}, we isolate language-related skills from the remaining set and categorise all ESCO-compliant skills into ``\textit{Hard Skills}'' and ``\textit{Soft Skills}'' following the ESCO classification scheme. 
The result is a harmonised and structured skill set for each dataset, consisting of three categories: ``\textit{Hard Skills}'', ``\textit{Soft Skills}'', and ``\textit{Language Skills}''.
Figure~\ref{fig:skill_preprocessing} provides a visual overview of the preprocessing pipeline for skills.

\paragraph{Working Hours and Contract Type.}

Job offer descriptions often include various contractual details.
Based on this observation, we designed a task for the LLM to extract two specific types of information: \textit{working hours} and \textit{contract type}.
For each job offer description, we issue two separate prompts to the LLM, each consisting of a query accompanied by the relevant context. 
The first prompt aims to classify the job as either a full-time or a part-time contract, based on the information provided in the job description. 
In cases where explicit references to working hours were absent, the LLM was instructed to infer the appropriate classification from the surrounding context.
The second one focused on identifying the nature of the contract duration: ``fixed-term'' or ``permanent''. 
This classification followed the same procedure as the working hours task, relying on both explicit cues and contextual inference when necessary.
This approach enabled the extraction of structured contractual information from unstructured job descriptions, contributing to a more comprehensive and standardised representation of job offers.
Finally, we emphasise that although we extracted the ``Contract Type'' feature from the job descriptions, we observed that it did not align with the curriculum characteristics.
Consequently, we opted not to include the feature in the experiments to preserve simplicity and interpretability in the results.

\end{document}